\documentclass{article}

\usepackage[main,final]{neurips_2026}

\usepackage[utf8]{inputenc} 
\usepackage[T1]{fontenc}    
\usepackage{hyperref}       
\hypersetup{hypertexnames=false,hidelinks}
\usepackage{url}            
\usepackage{booktabs}       
\usepackage{amsfonts}       
\usepackage{nicefrac}       
\usepackage{microtype}      
\usepackage[table]{xcolor}   
\usepackage{booktabs}
\usepackage{algorithm}
\usepackage{algpseudocode}
\usepackage{amsmath,amssymb}
\usepackage{graphicx}
\usepackage{multirow}
\usepackage{threeparttable}
\usepackage{array}
\usepackage[most]{tcolorbox}
\usepackage{upquote}
\usepackage{listings}
\usepackage{makecell}

\definecolor{oursblue}{HTML}{EAF4FF}       
\definecolor{llmorange}{HTML}{FFF3E0}      
\definecolor{heurgreen}{HTML}{EEF8EE}      
\definecolor{teacherpurple}{HTML}{F3ECFF}  
\definecolor{refgray}{HTML}{F2F2F2}        

\newcommand{\meanstd}[2]{\shortstack{#1\\{\scriptsize$\pm\,#2$}}}
\newcommand{\beststd}[2]{\shortstack{\textbf{#1}\\{\scriptsize$\pm\,#2$}}}
\newcolumntype{A}{>{\columncolor{oursblue}}c}
\newcolumntype{B}{>{\columncolor{llmorange}}c}
\newcolumntype{D}{>{\columncolor{heurgreen}}c}
\newcolumntype{G}{>{\columncolor{teacherpurple}}c}
\newcolumntype{K}{>{\columncolor{refgray}}c}
\title{Teacher-Aware Evolution of Heuristic Programs from Learned Optimization Policies}

\author{%
  Minyu Chen$^{1,4}$\thanks{Corresponding authors.} \quad Song Qin$^{1,3,4}$ \quad Ling-I Wu$^{2}$ \\
  \textbf{Jianxin Xue}$^{3}$ \quad \textbf{Guoqiang Li}$^{2,4}$\footnotemark[1] \\[0.4em]
  $^{1}$School of Artificial Intelligence, Shenzhen Technology University \\
  $^{2}$Shanghai Jiao Tong University \\
  $^{3}$Shanghai Polytechnic University \\
  $^{4}$AI Innovation and Application Joint Research Institute, \\
  Shenzhen Technology University \\
  \texttt{chenminyu@sztu.edu.cn} \quad \texttt{li.g@sjtu.edu.cn}
}

\newcommand{\tightparagraph}{\looseness=-1\relax}

\begin{document}

\maketitle

\begingroup
\renewcommand{\thefootnote}{\fnsymbol{footnote}}
\begin{abstract}
LLM-based automatic heuristic design has shown promise for generating executable heuristics for combinatorial optimization, but existing methods mainly rely on delayed endpoint performance. We propose a \emph{teacher-aware evolutionary framework} that uses independently trained learned optimization policies as behavioral teachers. During search, our method queries the teacher on states visited by candidate heuristic programs and uses its action preferences as local feedback for evolution. The resulting search discovers static executable heuristics guided by both task performance and teacher-derived behavioral signals. Experiments on scheduling, routing, and graph optimization benchmarks show that our method improves over performance-driven LLM heuristic evolution baselines while requiring no neural inference at deployment. These results suggest that learned optimization policies can be repurposed as behavioral feedback sources for automatic heuristic discovery.\footnote{Code and generated programs are available at \url{https://github.com/Minkow/Teacher-Aware-Evolution}.}
\end{abstract}
\endgroup

\section{Introduction}
\label{sec:intro}

Heuristics are indispensable in practical combinatorial optimization~\citep{bozorg2017meta,bello2017neural}. In scheduling, routing, graph optimization, and many other sequential decision problems, compact heuristic rules remain attractive because they are fast, interpretable, and easy to deploy~\citep{zhang2020learning}. Yet designing such rules is difficult. Strong heuristics often require domain expertise, repeated empirical tuning, and careful adaptation to each problem class.


Recent LLM-based automatic heuristic design (AHD) methods can generate executable heuristics through evaluation-driven search, evolutionary refinement, reflection, and population management~\citep{romeraparedes2024funsearch,liu2024eoh,ye2024reevo,yao2025meoh,liu2026eohs}. Yet their guidance remains largely internal to the LLM-evolution loop, relying on endpoint performance, textual feedback, or population-level statistics. 

In parallel, deep reinforcement learning (DRL) and neural combinatorial optimization (NCO) have produced learned policies for sequential combinatorial optimization~\citep{zhang2020learning,kwon2020pomo,luo2023neural,barrett2020exploratory}. Although these models are typically used as solvers, they are also rich sources of action-level behavioral information: given a state and a feasible action set, they provide preferences over candidate decisions. This information is complementary to endpoint fitness, but has not been used to guide LLM-based heuristic evolution.

We leverage such learned policies as \emph{behavioral teachers}. Rather than deploying or imitating the teacher, we query it on states induced by candidate heuristic programs and use its preferences as local search guidance. This is particularly useful when a learned policy is available during development, but deployment requires standalone heuristic code, as in CPU-only environments or applications requiring inspectable and editable decision rules. This on-policy querying is important: the teacher can evaluate decisions made along trajectories that differ from its own, enabling feedback on the actual states visited by evolving programs.

To this end, we introduce a teacher-aware evolutionary framework that turns local behavioral feedback into executable heuristic programs. Teacher-informed reflection guides structural rewrites, parameter calibration, and mechanism-level fusion, while multi-objective population management preserves complementary candidates during search. Task performance determines the final program, allowing evolution to exploit useful teacher preferences while departing from them when beneficial.

This perspective changes the role of reflection in LLM-based AHD. Prior methods mainly reflect on endpoint performance, population statistics, or previously generated programs~\citep{liu2024llm_evo,yao2025meoh,zheng2025mcts,wu2025efficient,vanstein2024llamea}. In contrast, our framework reflects on behavioral discrepancies between candidate programs and learned teachers. Such discrepancies indicate where a program's local decisions diverge from learned optimization behavior, and can guide the search toward heuristics that preserve useful learned biases while correcting teacher-specific weaknesses. The resulting output remains a static heuristic program, selected by task performance rather than teacher fidelity.

We evaluate our method on four sequential combinatorial optimization benchmarks: the job-shop scheduling problem (JSSP), the traveling salesman problem (TSP), the capacitated vehicle routing problem (CVRP), and the max-cut problem (MaxCut). These tasks cover scheduling, routing, and graph local search, and instantiate the same abstraction of selecting actions from feasible candidate sets under delayed endpoint feedback. Across these benchmarks, we improve over performance-driven LLM heuristic evolution baselines while producing static executable heuristics.
\par\noindent Our contributions are:
\begin{itemize}
    \item We formulate LLM-based heuristic evolution with independently trained black-box optimization policies as external behavioral teachers.
    \item We propose sampled on-policy teacher alignment, which queries the teacher on candidate-induced states and converts teacher preferences into local behavioral search signals.
    \item We introduce teacher-aware reflective evolution, using teacher-alignment summaries and disagreement cases to guide local refinement and structural revision of heuristic programs.
    \item We validate the framework on sequential combinatorial optimization tasks, showing that learned policies guide the discovery of static, interpretable, and efficient heuristics.
\end{itemize}

\section{Related Work}

\paragraph{LLM-based automatic heuristic design.}
Automatic heuristic design aims to reduce the human effort required to craft problem-specific heuristics and hyper-heuristics for combinatorial optimization~\citep{sabar2014automatic,stutzle2018automated,camacho2023designing,zhao2024automated,xi2025rise}. Recent LLM-based methods generate executable heuristic programs and improve them through evaluation-driven search. FunSearch~\citep{romeraparedes2024funsearch} showed that LLM-generated programs can be iteratively improved through empirical evaluation. EoH~\citep{liu2024eoh}, ReEvo~\citep{ye2024reevo}, and EoH-S~\citep{liu2026eohs} further develop evolutionary search, reflection, and population-level mechanisms for heuristic discovery. These methods primarily use feedback produced within the LLM search loop, such as endpoint objective values, textual reflections, and population statistics. Our work introduces an external source of behavioral feedback from independently trained learned optimization policies.

\paragraph{Learning-based policies for combinatorial optimization.}
Learning-based methods have been widely studied for combinatorial optimization, especially for sequential decision problems~\citep{bello2017neural,bengio2021mlco,li2022overview,mazyavkina2021rlco,darvariu2024grlco}. Examples include learning-to-dispatch policies for JSSP~\citep{zhang2020learning}, neural routing policies such as the Attention Model, POMO, and LEHD~\citep{kool2019attention,kwon2020pomo,luo2023neural}, and RL-based policies for packing and graph optimization~\citep{zhao2021online,pan2023adjustable,barrett2020exploratory}. These policies are typically used directly as solvers. In contrast, we use them as queryable behavioral teachers during heuristic evolution, extracting local preferences over feasible actions while outputting static heuristic programs.

\paragraph{Policy distillation and interpretable policy extraction.}
Policy distillation and interpretable policy extraction transfer a complex teacher policy into a simpler model or program~\citep{rusu2015policy,parisotto2017nsp,czarnecki2019distilling}. Interpretable RL further studies extraction into decision trees, symbolic expressions, or programmatic policies, such as VIPER~\citep{bastani2018viper} and PIRL~\citep{verma2018pirl}. Our goal is not faithful policy compression: the teacher may be imperfect, and direct imitation may not optimize the target objective. We instead use teacher preferences as auxiliary behavioral signals inside LLM-based evolutionary search, while selecting final heuristics by optimization performance.

\section{Method}
\label{sec:method}

\begin{figure}[!ht]
    \centering
    \includegraphics[width=\linewidth]{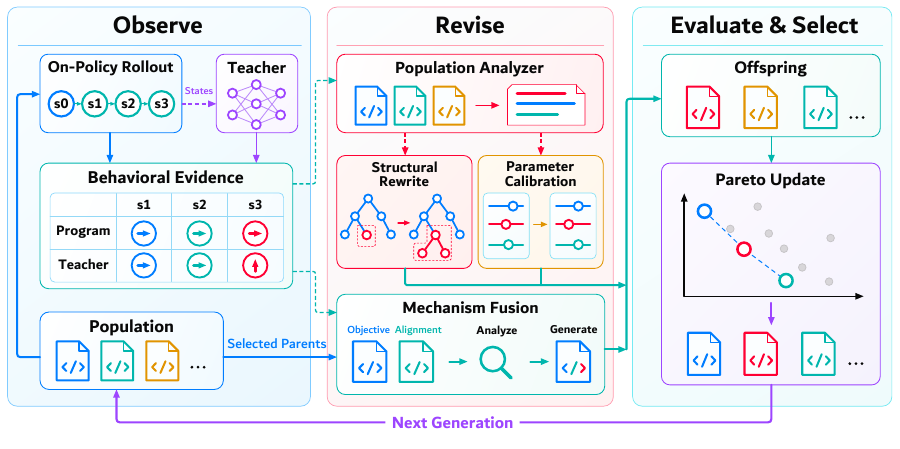}
    \caption{\small Teacher-aware heuristic evolution. Program and teacher decisions on visited states provide behavioral evidence. Population analysis guides rewrite and calibration; fusion analyzes its selected pair. Evaluated offspring undergo Pareto retention to form the next population.}
    \label{fig:framework}
\end{figure}

We propose a teacher-aware evolutionary framework for discovering executable heuristics for sequential combinatorial optimization. The framework follows recent LLM-based automatic heuristic design methods, where candidate programs are generated, evaluated, and improved through evolutionary search~\citep{liu2024eoh,ye2024reevo,liu2026eohs}. Our key addition is to use an independently trained black-box optimization policy as an external behavioral teacher. The teacher is queried on states induced by candidate programs, and its action preferences are used as auxiliary feedback for heuristic evolution rather than as imitation targets. Figure~\ref{fig:framework} summarizes the framework.

\subsection{Programmatic Sequential Heuristics}
\label{sec:programmatic_heuristics}

We consider a sequential decision problem over an instance \(x\). Solving \(x\) induces a trajectory of states \(s_0,\ldots,s_T\), where each state \(s_t\) has a feasible action set \(\mathcal A(s_t)\). A heuristic program \(h\) implements a deterministic action-selection rule:
\[
a_h(s_t)=h(s_t,\mathcal A(s_t)).
\]
A common special case is an executable scoring rule,
\[
a_h(s_t)=\arg\max_{a\in\mathcal A(s_t)}u_h(s_t,a),
\]
where \(u_h\) is implemented as code. After rollout, the program receives a task objective \(F(h;x)\). We write objectives in minimization form and seek
\[
\min_h \; \mathbb E_{x\sim\mathcal D}[F(h;x)].
\]
This abstraction covers both constructive heuristics, such as selecting the next operation or node, and local-search heuristics, such as selecting a move.

\subsection{On-policy Teacher Diagnostics}
\label{sec:teacher_diagnostics}

We assume query access to a learned teacher policy \(T\). Given a state \(s\) and feasible action set \(\mathcal A(s)\), the teacher returns either action scores or a preferred action. With action scores \(Q_T(s,a)\), we define
\[
a_T(s)=\arg\max_{a\in\mathcal A(s)}Q_T(s,a).
\]
Otherwise, \(a_T(s)\) is the teacher's returned action.

For each candidate \(h\), we roll it out and sample on-policy states \(\mathcal S_h\). The main alignment signal is top-1 agreement:
\[
\mathrm{Align}(h)
=
\frac{1}{|\mathcal S_h|}
\sum_{s\in\mathcal S_h}
\mathbf 1[a_h(s)=a_T(s)].
\]
\tightparagraph
We use top-1 agreement because it is comparable across teachers with different score scales and remains well-defined even when score calibration differs across domains. Rank-based correlations can be unstable when the feasible action set is small, while score-normalized values depend on teacher calibration.

When teacher scores are available, we min--max normalize them to $[0,1]$ over the feasible actions in each sampled state, assigning a value of 1 when all scores are equal. Averaging the normalized scores of the program's chosen actions over \(\mathcal S_h\) gives \(\mathrm{Value}(h)\); we also record their percentile ranks under the teacher ordering. These quantities are used as diagnostics and prompt features, while \(\mathrm{Align}(h)\) is the population-level alignment signal.

\subsection{Teacher-guided Reflection and Revision}
\label{sec:teacher_reflection}

The teacher diagnostics in Section~\ref{sec:teacher_diagnostics} provide action-level behavioral signals, but they must be converted into actionable feedback for program evolution. For each candidate program, we compare the action selected by the program with the teacher-preferred action on sampled on-policy states. These comparisons produce representative agreement and disagreement cases, together with score-based diagnostics when teacher values are available. Thus, the reflection signal is not only whether a program performs well, but also where its local decisions differ from the learned teacher.

\tightparagraph
Because the meaning of a disagreement depends on the task interface, we use a lightweight task interface adapter. The adapter specifies the program signature, objective convention, feasible-action abstraction, and diagnostic field names. Behavioral summaries are generated from candidate rollouts and teacher queries; the adapter defines their representation, and the LLM generates the rule.

A population analyzer is invoked once per teacher-active generation. It consumes population-level objective--alignment statistics, representative disagreement cases, and the task interface description, and produces concise revision briefs for structural rewrite and parameter calibration. These briefs connect behavioral mismatches to changes in the program's decision structure or numerical parameters. Mechanism-level fusion instead analyzes its selected pair and their behavioral evidence directly when generating an offspring. Table~\ref{tab:revision_modes} summarizes the three revision operators.

\paragraph{Structural rewrite.}
This mode modifies one higher-level decision component of a parent program to address recurring structural failures identified through teacher disagreements.

\paragraph{Parameter calibration.}
This mode preserves the program backbone and adjusts weights, thresholds, gates, phase boundaries, or tie-breaking priorities to correct local miscalibration identified through teacher feedback.

\paragraph{Mechanism-level fusion.}
Mechanism-level fusion analyzes two parents selected by task performance and teacher alignment, together with their rollout and teacher evidence. Using the better parent's performance as a reference, it recombines or transforms their decision mechanisms to improve the task objective.

\paragraph{Parent sampling.}
We use truncated rank-based sampling. Given a ranking criterion \(c\), let \(r_c(h)\) be the rank of program \(h\), with smaller rank better. The top-\(K\) pool is
\[
\mathcal P_K^{(c)}=\{h\in\mathcal P\mid r_c(h)\le K\}.
\]
A parent is sampled as
\[
p_c(h)=
\frac{w(r_c(h))}
{\sum_{h'\in\mathcal P_K^{(c)}}w(r_c(h'))},
\qquad h\in\mathcal P_K^{(c)},
\]
where \(w(\cdot)\) is a decreasing rank-weight function. Structural rewrite and parameter calibration sample from an objective-ranked pool; mechanism-level fusion samples one parent from an objective-ranked pool and one from an alignment-ranked pool.

\begin{table}[!ht]
\small
\setlength{\tabcolsep}{3pt}
\centering
\caption{\small Teacher-aware revision modes and the symbols used in the ablation study.}
\label{tab:revision_modes}
\footnotesize
\begin{tabular}{lll}
\toprule
Revision mode & Symbol & Main role \\
\midrule
Structural rewrite & \(s\) & modify one decision component \\
Parameter calibration & \(p\) & retune thresholds, weights, and gates \\
Mechanism-level fusion & \(m\) & \makecell[tl]{recombine or transform mechanisms\\using pair-local evidence} \\
\bottomrule
\end{tabular}
\end{table}

\subsection{Objective--Behavior Population Management}
\label{sec:population_update}

Population management is formulated as multi-objective selection. Let
\[
z(h)=\big(f_1(h),\ldots,f_d(h)\big)
\]
collect the criteria to be minimized. A candidate dominates another if it is no worse on every criterion and strictly better on at least one. Nondominated sorting preserves trade-offs among these criteria. Our experiments use three objectives: task performance, teacher alignment, and runtime,
\[
z(h)=\big(F(h),-\mathrm{Align}(h),R(h)\big),
\]
where \(R(h)\) is runtime per decision. At each update, current programs and eligible offspring are merged. Candidates within \(F_{\min}+\epsilon|F_{\min}|\) are considered first, adding candidates in objective order if needed to fill the population; nondominated sorting then retains programs over \(z(h)\), using crowding distance to preserve diversity when a front exceeds the remaining capacity. This follows the general multi-objective view of recent LLM-based heuristic evolution~\citep{yao2025meoh}, while incorporating teacher agreement as a behavioral criterion.

The final returned heuristic is selected by task objective.

\subsection{Overall Procedure}
\label{sec:overall_procedure}

At generation \(g\), each program in \(\mathcal P_g\) is rolled out on design instances to obtain its task objective, runtime, and sampled states. Teacher queries on these states produce top-1 agreement, normalized teacher value, percentile rank, and representative disagreement cases. This evidence supports population-level briefs for structural rewrite and parameter calibration, and pair-local reasoning for mechanism-level fusion. Each revision mode generates its allocated offspring, which are evaluated and merged with the current population. Multi-objective retention then produces \(\mathcal P_{g+1}\).

\subsection{Task Instantiations}
\label{sec:task_instantiations}

We instantiate the same action-selection interface across four tasks. Table \ref{tab:tasks} specifies the action space used by each program and the corresponding abstraction used for teacher alignment.

\begin{table}[!ht]
\centering

\setlength{\tabcolsep}{3pt}
\caption{\small Task instantiations. Each task is cast as selecting one action from a feasible candidate set under the current state.}
\label{tab:tasks}
\footnotesize
\begin{tabular}{llll}
\toprule
Task & Program action & Teacher & Alignment action \\
\midrule
JSSP & dispatch operation & Offline-L2D & operation \\
TSP & choose next node & LEHD & node \\
CVRP & choose customer / restart route & LEHD & customer / restart \\
MaxCut & flip vertex & ECO-DQN & vertex flip \\
\bottomrule
\end{tabular}
\end{table}

\paragraph{JSSP.}
For the job-shop scheduling problem (JSSP), the feasible action set consists of operations that can be dispatched at the current state. A heuristic program selects one operation to schedule next, and teacher alignment is measured over the same feasible operation set.

\paragraph{TSP.}
For the traveling salesman problem (TSP), the heuristic constructs a tour by selecting the next unvisited node. Teacher alignment is computed over the same unvisited-node candidate set.

\paragraph{CVRP.}
For the capacitated vehicle routing problem (CVRP), the abstract action set contains all currently feasible customers and a depot-restart action. Returning the depot closes the current route; the next decision starts from the depot with reset capacity. When the teacher exposes better route-continuation actions, we fold them into the same abstract action space used by the heuristic.

\paragraph{MaxCut.}
For the max-cut problem (MaxCut), we use a flip-based local-search formulation. The heuristic directly returns a vertex to flip, and all vertices are feasible actions. Flip history is provided as state information and may be used by the generated heuristic as a soft anti-cycling signal, but it is not a hard action mask. Teacher alignment is computed over the same vertex-flip action set.




\section{Experiments}
\subsection{Experimental Setup}
\label{sec:experimental_setup}

\paragraph{Benchmarks and evaluation protocol.}
We evaluate on four sequential combinatorial optimization benchmarks: JSSP, TSP, CVRP, and MaxCut. These tasks cover scheduling, routing, and graph local search, and all can be formulated as repeatedly selecting one action from a feasible candidate set under delayed endpoint feedback. 

For each task, heuristic programs are evolved on a fixed design set and then evaluated on both in-distribution and generalization benchmarks. Generalization benchmarks differ from the design setting in scale, distribution, or benchmark source. For example, routing heuristics evolved on TSP50/CVRP50 are evaluated on larger synthetic instances and standard benchmark instances such as TSPLIB/VRPLIB.

\paragraph{Teachers and baselines.}
Each task uses an independently trained optimization policy as the behavioral teacher. We use an offline Learning to Dispatch (L2D)-style policy for JSSP~\citep{zhang2020learning,vanremmerden2025offline}, Light Encoder and Heavy Decoder (LEHD) for TSP and CVRP~\citep{luo2023neural}, and Exploratory Combinatorial Optimization with Deep Q-Networks (ECO-DQN) for MaxCut~\citep{barrett2020exploratory}. Teacher policies are queried only during heuristic evolution; their direct rollout performance is reported as a reference.

We compare against performance-driven LLM heuristic evolution baselines, including EoH-style evolutionary search~\citep{liu2024eoh} and ReEvo-style reflective evolution~\citep{ye2024reevo}. We also include classical heuristics and strong solvers or reference solutions when applicable. All LLM-based methods use the same initial seed programs and design instances.

\paragraph{LLM budget and implementation.}
LLM-based methods use GPT-5.4, except the MaxCut EoH OOD results and the selection-alignment ablation, which use GPT-5.6 Terra with medium reasoning. Each run has a budget of 100 FE. Our method uses population size 12 and five generations. We uniformly sample 20 states per sampled trajectory. Detailed allocations and task settings appear in Appendix~\ref{app:configuration_provenance}.

\paragraph{Teacher interface.}
We use released public checkpoints without retraining or fine-tuning. The teacher accepts task-level features and returns a preferred action or scores over feasible alternatives through a black-box query interface. Alignment measures agreement on each candidate's own induced states, using common design instances and a uniform query budget across candidates.

\subsection{Main Results}
\label{sec:main_results}

Teacher-aware evolution achieves lower makespan and routing cost than the performance-driven LLM baselines across the reported JSSP, TSP, and CVRP settings, including larger instances and standard benchmark sets. On MaxCut, it achieves the highest mean cut on all seven graph sets.

\subsubsection{Results on the Job-Shop Scheduling Problem}
\label{sec:jssp_results}

We first evaluate on JSSP, where a heuristic repeatedly selects one feasible operation to dispatch and the final objective is makespan. We compare against LLM-based heuristic evolution baselines, classical dispatching rules, and learned dispatching policies.

Table~\ref{tab:jssp_random} reports results on randomly generated JSSP instances. Our method achieves the best makespan across \(10\times10\), \(15\times15\), and \(20\times20\), outperforming EoH, ReEvo, classical dispatching rules, and learned dispatching policies. The timing evaluation shows faster execution for generated programs than for neural policies, particularly Offline-LD.

\begin{table}[!ht]
\centering
\caption{\small Results on randomly generated JSSP instances. We report makespan and runtime separately. Lower is better for both metrics. Column colors indicate method families: blue for our method, orange for LLM-based heuristic evolution baselines, green for classical dispatching rules, and purple for learned dispatching policies.}
\label{tab:jssp_random}
\footnotesize
\setlength{\tabcolsep}{2.2pt}
\renewcommand{\arraystretch}{1.18}
\setlength{\extrarowheight}{1pt}
\renewcommand{\meanstd}[2]{\ensuremath{\vcenter{\hbox{\shortstack{#1\\{\scriptsize$\pm\,#2$}}}}}}
\renewcommand{\beststd}[2]{\ensuremath{\vcenter{\hbox{\shortstack{\textbf{#1}\\{\scriptsize$\pm\,#2$}}}}}}
\begin{tabular}{llA B B D D D D G G}
\toprule
Dataset & Metric & Ours & EoH & ReEvo & SPT & MWKR & \makecell{FDD-\\MWKR} & MOR & L2D & \makecell{Offline-\\LD} \\
\midrule
\multirow{2}{*}{\(10\times10\)}
& Makespan \(\downarrow\)
& \beststd{913.138}{4.470} & \meanstd{925.444}{11.115} & \meanstd{921.360}{8.135} & 1231.49 & 990.26 & 955.38 & 983.91 & 988.26 & 976.52 \\
& Time (s) \(\downarrow\)
& 1.81 & 1.44 & 1.51 & 2.69 & 2.85 & 2.85 & 2.84 & 17.45 & 12.51\rule[-6pt]{0pt}{13pt} \\

\multirow{2}{*}{\(15\times15\)}
& Makespan \(\downarrow\)
& \beststd{1372.040}{8.601} & \meanstd{1393.182}{13.187} & \meanstd{1377.508}{11.508} & 1879.15 & 1504.41 & 1463.33 & 1497.08 & 1508.62 & 1499.30 \\
& Time (s) \(\downarrow\)
& 6.06 & 6.75 & 5.78 & 8.66 & 9.77 & 9.87 & 9.73 & 37.01 & 114.26\rule[-6pt]{0pt}{13pt} \\

\multirow{2}{*}{\(20\times20\)}
& Makespan \(\downarrow\)
& \beststd{1812.160}{19.627} & \meanstd{1821.652}{30.064} & \meanstd{1838.220}{18.423} & 2542.51 & 1976.06 & 1946.20 & 1994.22 & 1998.63 & 1971.33 \\
& Time (s) \(\downarrow\)
& 18.59 & 20.80 & 18.41 & 23.09 & 27.09 & 27.37 & 27.02 & 68.10 & 228.69\rule[-5pt]{0pt}{12pt} \\
\bottomrule
\end{tabular}
\end{table}

Table~\ref{tab:jssp_ood} evaluates out-of-distribution generalization. All learned heuristic programs are evolved on \(20\times20\) instances and transferred to larger random instances and Taillard benchmarks. Our method achieves the best makespan on all OOD settings, suggesting that teacher-aware evolution discovers dispatching rules that transfer beyond the design scale and distribution.

\begin{table}[!ht]
\centering
\caption{\small OOD generalization on JSSP. Heuristics evolved on \(20\times20\) are evaluated on larger random instances and Taillard benchmark instances. We report makespan and runtime separately. Lower is better for both metrics. Column colors indicate method families: blue for our method, orange for LLM-based heuristic evolution baselines, green for classical dispatching rules, and purple for learned dispatching policies.}
\label{tab:jssp_ood}
\footnotesize
\setlength{\tabcolsep}{2.2pt}
\renewcommand{\arraystretch}{1.18}
\setlength{\extrarowheight}{1pt}
\renewcommand{\meanstd}[2]{\ensuremath{\vcenter{\hbox{\shortstack{#1\\{\scriptsize$\pm\,#2$}}}}}}
\renewcommand{\beststd}[2]{\ensuremath{\vcenter{\hbox{\shortstack{\textbf{#1}\\{\scriptsize$\pm\,#2$}}}}}}
\begin{tabular}{llA B B D D D D G G}
\toprule
Dataset & Metric
& Ours & EoH & ReEvo
& SPT & MWKR & \makecell{FDD-\\MWKR} & MOR
& L2D & \makecell{Offline-\\LD} \\
\midrule

\multirow{2}{*}{\makecell{Random\\\(30\times20\)}}
& Makespan \(\downarrow\)
& \beststd{2279.572}{25.136} & \meanstd{2297.344}{40.051} & \meanstd{2315.582}{25.997}
& 3222.13 & 2499.05 & 2454.21 & 2483.91
& 2481.87 & 2476.46 \\
& Time (s) \(\downarrow\)
& 56.25 & 63.92 & 55.34
& 61.07 & 75.25 & 76.19 & 75.19
& 113.48 & 455.81\rule[-6pt]{0pt}{13pt} \\

\multirow{2}{*}{\makecell{Random\\\(50\times20\)}}
& Makespan \(\downarrow\)
& \beststd{3231.568}{26.680} & \meanstd{3274.420}{47.843} & \meanstd{3293.656}{35.158}
& 4486.27 & 3513.67 & 3467.97 & 3489.32
& 3457.56 & 3476.96 \\
& Time (s) \(\downarrow\)
& 234.48 & 274.93 & 236.04
& 229.91 & 292.70 & 294.66 & 293.36
& 228.95 & 1022.98\rule[-6pt]{0pt}{13pt} \\

\multirow{2}{*}{\makecell{Taillard\\\(20\times20\)}}
& Makespan \(\downarrow\)
& \beststd{1913.980}{29.121} & \meanstd{1923.320}{30.052} & \meanstd{1940.820}{18.991}
& 2672.40 & 2079.50 & 2015.40 & 2069.70
& 2097.30 & 2071.10 \\
& Time (s) \(\downarrow\)
& 1.83 & 2.07 & 1.78
& 2.43 & 2.70 & 2.74 & 2.70
& 10.30 & 26.96\rule[-6pt]{0pt}{13pt} \\

\multirow{2}{*}{\makecell{Taillard\\\(30\times20\)}}
& Makespan \(\downarrow\)
& \beststd{2377.680}{32.442} & \meanstd{2409.820}{35.677} & \meanstd{2425.880}{30.322}
& 3260.40 & 2615.00 & 2565.70 & 2619.80
& 2598.70 & 2607.00 \\
& Time (s) \(\downarrow\)
& 6.40 & 6.35 & 6.32
& 6.11 & 7.43 & 7.52 & 7.44
& 14.11 & 41.95\rule[-5pt]{0pt}{12pt} \\
\bottomrule
\end{tabular}
\end{table}

\subsubsection{Results on Routing Problems}
\label{sec:routing_results}

We next evaluate on routing problems. Heuristics are evolved on Euclidean TSP50 and CVRP50 design instances, and then evaluated on both in-distribution and generalization settings. For TSP, we evaluate on TSP50, TSP200, and TSPLIB instances. For CVRP, we evaluate on CVRP50, CVRP200, and VRPLIB instances. We compare against LEHD as a learned routing teacher, and LKH or reference solutions as strong optimization references.

\tightparagraph
Tables~\ref{tab:tsp_results} and~\ref{tab:cvrp_results} show that our method consistently obtains the best solution quality among LLM-evolved heuristics. On TSP, our method improves over both EoH and ReEvo across in-distribution and OOD settings, including TSP200 and TSPLIB. On CVRP, our method also achieves the best LLM-evolved heuristic performance on all settings and transfers to larger CVRP200 and VRPLIB instances. The gains on 200-node instances demonstrate transfer beyond the 50-node design scale, while the TSPLIB and VRPLIB results extend this evidence to distinct benchmark sources.

\begin{table}[!ht]
\centering
\caption{\small Results on TSP. Heuristics are evolved on TSP50 and evaluated on in-distribution and generalization settings. Lower is better for both tour length and runtime. Column colors indicate method families: blue for our method, orange for LLM-based heuristic evolution baselines, purple for learned neural policies, and gray for strong solver/reference results.}
\label{tab:tsp_results}
\footnotesize
\setlength{\tabcolsep}{2pt}
\renewcommand{\arraystretch}{1.12}
\setlength{\extrarowheight}{0pt}
\renewcommand{\meanstd}[2]{\ensuremath{#1\,{\scriptstyle\pm\,#2}}}
\renewcommand{\beststd}[2]{\ensuremath{\mathbf{#1}\,{\scriptstyle\pm\,#2}}}
\begin{threeparttable}
\begin{tabular}{llA B B G K}
\toprule
Dataset & Metric & Ours & EoH & ReEvo & LEHD & LKH \\
\midrule
\multirow{2}{*}{TSP50}
& Tour length \(\downarrow\)
& \beststd{6.498}{0.061} & \meanstd{6.545}{0.156} & \meanstd{6.599}{0.074} & 5.739 & 5.709 \\
& Time (s) \(\downarrow\)
& 0.28 & 0.13 & 0.05 & 0.10 & 1.05 \\

\multirow{2}{*}{TSP200}
& Tour length \(\downarrow\)
& \beststd{12.499}{0.127} & \meanstd{12.713}{0.231} & \meanstd{12.947}{0.204} & 10.796 & 10.70 \\
& Time (s) \(\downarrow\)
& 4.49 & 0.77 & 0.14 & 2.97 & 11.75 \\

\multirow{2}{*}{TSPLIB70}
& Tour length \(\downarrow\)
& \beststd{87969.551}{1111.993} & \meanstd{89535.581}{1756.125} & \meanstd{90326.130}{1378.918} & 78511.262 & 73156.01 \\
& Time (s) \(\downarrow\)
& 464.92 & 505.64 & 6.66 & 388.81 & \(>1\)h \\
\bottomrule
\end{tabular}
\end{threeparttable}
\end{table}

\begin{table}[!ht]
\centering
\caption{\small Results on CVRP. Heuristics are evolved on CVRP50 and evaluated on in-distribution and generalization settings. Lower is better for both route length and runtime. Column colors indicate method families: blue for our method, orange for LLM-based heuristic evolution baselines, purple for learned neural policies, and gray for strong solver/reference results.}
\label{tab:cvrp_results}
\footnotesize
\setlength{\tabcolsep}{2pt}
\renewcommand{\arraystretch}{1.12}
\setlength{\extrarowheight}{0pt}
\renewcommand{\meanstd}[2]{\ensuremath{#1\,{\scriptstyle\pm\,#2}}}
\renewcommand{\beststd}[2]{\ensuremath{\mathbf{#1}\,{\scriptstyle\pm\,#2}}}
\begin{tabular}{llA B B G K}
\toprule
Dataset & Metric & Ours & EoH & ReEvo & LEHD & LKH \\
\midrule
\multirow{2}{*}{CVRP50}
& Route length \(\downarrow\)
& \beststd{9.650}{0.182} & \meanstd{9.865}{0.162} & \meanstd{9.720}{0.197} & 8.180 & 7.828 \\
& Time (s) \(\downarrow\)
& 0.16 & 1.09 & 0.45 & 0.11 & \(>1\)h \\

\multirow{2}{*}{CVRP200}
& Route length \(\downarrow\)
& \beststd{25.2165}{0.3829} & \meanstd{25.5913}{0.4558} & \meanstd{26.5206}{0.7272} & 20.832 & 20.34 \\
& Time (s) \(\downarrow\)
& 7.12 & 59.20 & 22.89 & 2.65 & \(>1\)h \\

\multirow{2}{*}{VRPLIB192}
& Route length \(\downarrow\)
& \beststd{38099.154}{936.703} & \meanstd{38740.827}{1704.914} & \meanstd{39727.084}{2020.635} & 39120.68 & 33281.32 \\
& Time (s) \(\downarrow\)
& 282.31 & 1466.31 & 536.32 & 234.03 & \(>1\)h \\
\bottomrule
\end{tabular}
\end{table}

\subsubsection{Results on the Max-Cut Problem}
\label{sec:maxcut_results}

We further evaluate graph local-search generalization on MaxCut. Heuristic programs are evolved on weighted Barabási--Albert graphs with 200 vertices (BA200w) and transferred to graphs with different distributions, sizes, and weighting schemes. Here, BA and ER denote Barabási--Albert and Erd\H{o}s--Rényi graphs, respectively; \(w\) and \(u\) denote weighted and unweighted graphs. We also evaluate on Physics graphs as an additional real-world graph benchmark. We compare against EoH and ReEvo as performance-driven LLM heuristic evolution baselines, and ECO-DQN as the learned policy reference.

Table~\ref{tab:maxcut_results} shows that our method achieves higher mean cuts than both EoH and ReEvo on BA200w and all six OOD graph sets. The gain on ER200w demonstrates transfer from BA to ER topology at the same graph size. The advantage persists on 800-vertex graphs with weighted and unweighted edges, supporting transfer across scale and weighting. The gains on Physics extend this evidence beyond synthetic graph families.

\begin{table}[!ht]
\centering
\caption{\small MaxCut quality and running time. BA200w is in-distribution; all remaining graph sets are OOD. Ours, EoH, and ReEvo use 16 CPU workers; times are measured over each complete test set. ECO-DQN uses the official dataset-specific policy and reference timings.}
\label{tab:maxcut_results}
\footnotesize
\setlength{\tabcolsep}{3.5pt}
\renewcommand{\arraystretch}{1.12}
\renewcommand{\meanstd}[2]{\ensuremath{#1\,{\scriptstyle\pm\,#2}}}
\renewcommand{\beststd}[2]{\ensuremath{\mathbf{#1}\,{\scriptstyle\pm\,#2}}}
\begin{tabular}{llA B B G}
\toprule
Dataset & Metric & Ours & EoH & ReEvo & ECO-DQN \\
\midrule
\multirow{2}{*}{BA200w} & Cut $\uparrow$ & \beststd{189.358}{0.921} & \meanstd{188.932}{0.512} & \meanstd{187.932}{0.435} & 184.32 \\
& Time (s) $\downarrow$ & 31.18 & -- & 19.19 & 126.19 \\
\midrule
\multirow{2}{*}{ER200w} & Cut $\uparrow$ & \beststd{397.276}{2.294} & \meanstd{394.654}{5.019} & \meanstd{394.088}{3.064} & 394.39 \\
& Time (s) $\downarrow$ & 34.87 & 67.59 & 21.51 & 134.77 \\
\multirow{2}{*}{BA800u} & Cut $\uparrow$ & \beststd{2297.544}{7.896} & \meanstd{2280.830}{8.796} & \meanstd{2281.554}{15.033} & 2286.85 \\
& Time (s) $\downarrow$ & 529.94 & 982.76 & 285.25 & 135.26 \\
\multirow{2}{*}{BA800w} & Cut $\uparrow$ & \beststd{718.004}{11.289} & \meanstd{700.716}{10.013} & \meanstd{696.992}{10.001} & 703.43 \\
& Time (s) $\downarrow$ & 471.13 & 954.09 & 283.04 & 134.31 \\
\multirow{2}{*}{ER800u} & Cut $\uparrow$ & \beststd{11334.720}{7.517} & \meanstd{11333.440}{30.193} & \meanstd{11333.040}{22.795} & 11325.80 \\
& Time (s) $\downarrow$ & 97.37 & 153.17 & 46.02 & 55.42 \\
\multirow{2}{*}{ER800w} & Cut $\uparrow$ & \beststd{1860.760}{14.010} & \meanstd{1854.240}{11.067} & \meanstd{1847.560}{24.286} & 1839.00 \\
& Time (s) $\downarrow$ & 72.60 & 136.90 & 46.04 & 58.27 \\
\midrule
\multirow{2}{*}{Physics} & Cut $\uparrow$ & \beststd{109.600}{0.510} & \meanstd{109.320}{0.482} & \meanstd{108.640}{1.053} & 108.40 \\
& Time (s) $\downarrow$ & 1.92 & 3.90 & 1.23 & 18.88 \\
\bottomrule
\end{tabular}
\end{table}

\subsection{Ablation Study}

\label{sec:matched_ablation}

\tightparagraph
We evaluate the contribution of teacher feedback and the three revision modes on JSSP20 and TSP50 (Table~\ref{tab:ablation}). Performance-only (EoH) provides the objective-driven evolutionary baseline. We separately remove structural rewrite ($s$), parameter calibration ($p$), or mechanism fusion ($m$), dividing its offspring budget equally between the other two modes to maintain 100 FE. The teacher-signal ablation removes teacher feedback while retaining the operator structure. The selection-alignment ablation retains objective and runtime criteria, teacher feedback, and alignment-based parent sampling.

\begin{table}[!ht]
\centering
\caption{\small Ablation study at 100 FE. Lower is better. The last TSP50 entry summarizes four runs, excluding the run with tour length 5.909; all five runs yield $6.414\pm0.309$.}
\label{tab:ablation}
\label{tab:ablation_matched}
\small
\renewcommand{\arraystretch}{1.25}
\setlength{\tabcolsep}{14pt}
\newcommand{\ablstd}[1]{{\footnotesize\textcolor{black!65}{#1}}}
\begin{tabular}{@{}l r@{\,$\pm$\,}l r@{\,$\pm$\,}l@{}}
\toprule
\textbf{Variant} & \multicolumn{2}{c}{\textbf{JSSP20} $\downarrow$} & \multicolumn{2}{c@{}}{\textbf{TSP50} $\downarrow$} \\
\midrule
\textbf{Full} & 1812.160 & \ablstd{19.627} & 6.498 & \ablstd{0.061} \\
Performance-only (EoH) & 1821.652 & \ablstd{30.064} & 6.545 & \ablstd{0.156} \\
\midrule
w/o structural rewrite ($s$) & 1842.670 & \ablstd{21.580} & 6.575 & \ablstd{0.158} \\
w/o parameter calibration ($p$) & 1818.140 & \ablstd{29.200} & 6.527 & \ablstd{0.105} \\
w/o mechanism fusion ($m$) & 1837.980 & \ablstd{13.430} & 6.599 & \ablstd{0.074} \\
\midrule
w/o teacher signal & 1848.670 & \ablstd{15.620} & 6.612 & \ablstd{0.103} \\
w/o selection alignment & 1827.174 & \ablstd{14.489} & 6.540 & \ablstd{0.146} \\
\bottomrule
\end{tabular}
\end{table}

\tightparagraph
Among the operator and teacher-feedback ablations, Full has the lowest mean on both tasks. Removing parameter calibration produces the smallest gap (5.98 makespan on JSSP20 and 0.029 tour length on TSP50). Removing structural rewrite or mechanism fusion has a larger effect on both tasks, suggesting that changes to the decision-rule structure are particularly important within the fixed search budget. Removing teacher feedback increases the means by 36.51 and 0.114, respectively. These JSSP20 and TSP50 results support combining the three revision modes with behavioral guidance under a fixed search budget. Removing selection alignment increases the mean makespan by 15.01 on JSSP20. On TSP50, the four-run subset has a mean tour length 0.042 above Full and a larger standard deviation (0.146 versus 0.061), while the complete five-run result has a lower mean than Full.

The two teacher-related ablations distinguish guidance during program revision from alignment during population selection. Removing teacher feedback worsens the mean objective on both tasks, whereas removing selection alignment has a task-dependent effect. This contrast suggests that behavioral diagnosis and alignment-based retention play different roles in the search, rather than providing interchangeable sources of guidance.


\subsection{Search Dynamics}
\label{sec:search_dynamics}

\begin{figure}[!ht]
    \centering
    \includegraphics[width=0.98\linewidth]{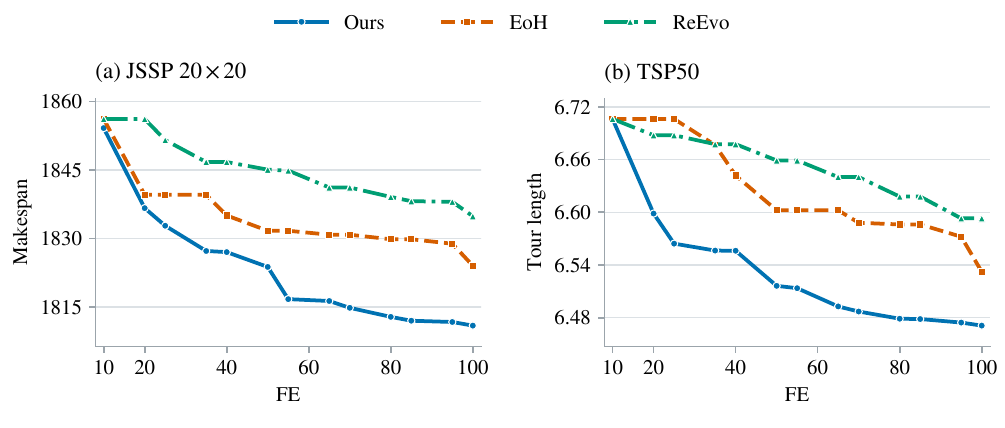}
    \caption{Mean best-so-far training objective over five runs at matched FE checkpoints.}
    \label{fig:learning_dynamics}
\end{figure}

Figure~\ref{fig:learning_dynamics} compares the mean best-so-far training objectives of our method, EoH, and ReEvo on JSSP \(20\times20\) and TSP50. Our method reaches lower mean training objectives than both baselines by 20 FE on both tasks and retains this advantage through 100 FE, indicating an early and sustained improvement in search efficiency.

\section{Conclusion and Limitations}
\label{sec:conclusion}

We introduce a teacher-aware evolutionary framework for automatic heuristic design. The method queries learned optimization policies on states visited by candidate programs and uses their action preferences as behavioral feedback, rather than deploying or imitating them directly. This enables LLM-based evolution to discover static executable heuristics guided by local decision signals beyond delayed endpoint objectives. Experiments show consistent improvements over performance-driven LLM heuristic evolution baselines, with no neural inference required at deployment. The framework requires a learned teacher with a compatible action interface, task-specific adapters, and analyzer calls during search. Its effectiveness also depends on the quality of teacher feedback. Future work may automate interface construction and invoke teacher feedback adaptively.

\bibliographystyle{plainnat}
\bibliography{refs}


\clearpage
\appendix

\section{Algorithm}
Algorithm~\ref{alg:overall_procedure} gives the corresponding procedure. The teacher is used only to provide behavioral feedback during search; the returned heuristic is selected by task objective and does not require neural inference at deployment.

\begin{algorithm}[!ht]
\caption{Teacher-Aware Evolution of Heuristic Programs}
\label{alg:overall_procedure}
\begin{algorithmic}[1]
\Require Initial population \(\mathcal P_0\), design instances \(\mathcal D\), teacher policy \(T\), population size \(N\), generations \(G\)
\Ensure Best heuristic program selected by task objective
\For{\(g=0,\ldots,G-1\)}
    \State Evaluate each \(h\in\mathcal P_g\) on \(\mathcal D\) to obtain \(F(h)\) and runtime \(R(h)\)
    \State Sample on-policy states \(\mathcal S_h\) from rollouts of each \(h\)
    \State Query teacher \(T\) on \(\mathcal S_h\) and compute \(\mathrm{Align}(h)\) and auxiliary diagnostics
    \State Construct representative disagreement cases from program actions and teacher-preferred actions
    \State Invoke the population analyzer for structural rewrite and parameter calibration
    \State Analyze each fusion pair within its offspring-generation request
    \State Generate offspring programs \(\mathcal C_g\) using teacher-aware revision modes
    \State Evaluate valid offspring programs and compute their teacher diagnostics
    \State Merge populations: \(\mathcal R_g \gets \mathcal P_g \cup \mathcal C_g\)
    \State Retain next population: \(\mathcal P_{g+1}\gets \textsc{ParetoRetain}(\mathcal R_g,N;F,\mathrm{Align},R)\)
\EndFor
\State \Return program in \(\mathcal P_G\) with the best task objective
\end{algorithmic}
\end{algorithm}

\section{Experimental and Runtime Settings}
\label{app:runtime_settings}

Each evolutionary-search run uses 16 Intel Xeon Gold 6336Y CPU cores at 2.40 GHz, one NVIDIA A10 GPU, and 256 GB memory. The software environment uses Python 3.11.15, PyTorch 2.8.0 with CUDA 12.8, and NumPy 2.4.4.

The five-run selection-alignment ablations took approximately 29--66 minutes per search on JSSP20 and 38--53 minutes on TSP50; the five MaxCut EoH searches took 107--166 minutes each. These 15 searches totaled approximately 19 job-hours, corresponding to 300 allocated CPU-core-hours and 19 allocated GPU-hours, including time spent waiting for LLM responses.

Generated heuristics are evaluated on CPUs, while learned policies use GPU inference in their corresponding implementations. Reported running times measure wall-clock evaluation time; task-specific execution settings are given with the results.

\section{Experimental Details}
\label{app:configuration_provenance}

\paragraph{Task-specific diversity controls.}
For JSSP10, JSSP20, and CVRP50, repeated fusion offspring from the same unordered parent pair are restricted to one objective-best eligible child per generation. For JSSP15, Max-Cut, and TSP, fusion pairs are sampled without replacement, with at most two children per performance-reference parent in each batch. In TSP, pair-local diagnosis is performed within the fusion request, independently of the population analyzer. Parameter-calibration children sharing an immediate parent are restricted to one objective-best eligible child per generation. All evaluated children count toward the FE budget.

\paragraph{Search allocation and selection.}
Five shared seed programs and five initialization programs are followed by schedules $(8,8,5,5,4)$ for each of the three revision modes, totaling 100 FE. The first eight structural-rewrite children use the top three objective-ranked, code-distinct parents with allocation $3/3/2$. The main population size is 12. Objective slack is 2\% for JSSP and Max-Cut and 5\% for TSP and CVRP. Runtime is an auxiliary Pareto criterion.

\paragraph{TSP cohort selection.}
The TSP results summarize five selected programs, with seed 3 replaced by a rerun after observing the original completed result. Within each run, the training objective selects the program. The same five programs are evaluated on all TSP test sets.

\paragraph{Runtime measurement.}
\tightparagraph
Both generated heuristics and LEHD make dependent sequential decisions within an instance. Generated programs evaluate independent instances with 16 Intel Xeon Gold 6336Y CPU workers, while the official LEHD implementation batches independent instances on an RTX 3090. LKH uses its official CPU implementation. Per-instance latency measures individual solve time; parallel wall time measures batch throughput. Teacher and LLM calls occur during search.

\paragraph{Beyond constructive heuristics.}
The sequential-action interface also accommodates local search, as illustrated by vertex-flip decisions in Max-Cut. With a suitable queryable teacher, iterative local search could expose move, acceptance, perturbation, or restart decisions; an evolutionary algorithm could expose parent, variation-operator, or survivor selection. LNS could expose destroy/repair decisions, and branch-and-cut could expose branching or cut selection.

\paragraph{Relation to imitation-only controls.}
Policy distillation targets teacher fidelity, whereas our method optimizes task performance and may depart from an imperfect teacher. The teacher-signal ablation in Table~\ref{tab:ablation} tests the contribution of behavioral feedback to evolutionary search.

\paragraph{Code availability.}
The implementation, generated programs, prompts, configurations, and evaluation instructions are available in our \href{https://github.com/Minkow/Teacher-Aware-Evolution}{GitHub repository}. We credit the original sources and follow the licenses and terms of use of the datasets, code, and pretrained models.

\section{Broader Impacts}
\label{app:broader_impacts}
\tightparagraph
Inspectable scheduling, routing, and graph programs can be easier to audit, adapt, and deploy on CPUs. Benchmark objectives may omit fairness, service quality, worker constraints, or other application requirements. Generated code may contain defects or fail under distribution shift, and LLM search consumes computational resources. Deployment requires code review, feasibility checks, representative operational testing, and explicit application constraints.


\section{Task Interface Adapter Prompts}
\label{app:task_interface_prompts}

The task interface adapter specifies the executable function signature, valid inputs and outputs, action semantics, and diagnostic fields exposed to the LLM. It ensures that generated programs follow the task interface and expresses teacher-derived behavioral summaries in task-specific terms. The LLM supplies the heuristic's action-ranking logic.

All generation prompts follow the same output contract: the LLM first gives a one-sentence algorithm description enclosed in braces, and then implements the required Python function. No additional prose is allowed outside the requested description and code.

\paragraph{JSSP.}
The JSSP interface asks the LLM to implement a priority scoring rule:
\begin{verbatim}
Function name: score
Inputs: feature, state
Output: priority_score
\end{verbatim}
The scheduler evaluates this function on every feasible operation and dispatches the operation with the highest score. The input \texttt{feature} contains operation-level attributes such as processing time, remaining work, machine id, earliest start, earliest finish, machine queue length, machine load, job progress, and lower-bound information. The function must return a scalar priority score.

The state is a partial schedule, and each unfinished job contributes its next unscheduled operation to the feasible set. Dispatching an operation inserts it at the earliest feasible position on its machine, respecting job precedence and machine occupancy. The rollout ends when all operations are scheduled, and its objective is the makespan. Offline-L2D preferences are mapped to the same feasible operation identifiers.

\paragraph{TSP.}
The TSP interface asks the LLM to implement a next-node selector:
\begin{verbatim}
Function name: select_next_node
Inputs: current_node, destination_node, unvisited_nodes, distance_matrix
Output: next_node
\end{verbatim}
The function returns one node id from the unvisited set, or the destination node when no unvisited node remains.

The state consists of the current path endpoint, the unvisited nodes, and the distance matrix. Selecting a node extends the path and removes that node from the unvisited set; the final connection closes the tour. The objective is total tour length. LEHD's candidate-node preferences are mapped back to the node identifiers used by the heuristic.

\paragraph{CVRP.}
The CVRP interface extends the routing selector with capacity constraints:
\begin{verbatim}
Function name: select_next_node
Inputs: current_node, depot, unvisited_nodes,
        rest_capacity, demands, distance_matrix
Output: next_node
\end{verbatim}
The function returns either a feasible customer whose demand does not exceed the remaining capacity, or the depot id \(0\) to restart the route.

The state tracks the current location, unserved customers, and remaining vehicle capacity. Visiting a customer removes it from the unserved set and deducts its demand; returning to the depot resets capacity. The rollout ends after every customer is served and the final route returns to the depot, with total route length as the objective. Direct-customer teacher actions map to customer identifiers, while via-depot actions map to the restart action; the restart score is the maximum teacher score among these via-depot choices.

\paragraph{MaxCut.}
The MaxCut interface asks the LLM to implement a deterministic flip selector:
\begin{verbatim}
Function name: select_flip_node
Inputs: spins, adjacency, current_cut, best_cut,
        step, max_steps, flip_history
Output: node_id
\end{verbatim}
The function returns an integer vertex id in \([0,n-1]\). All vertices are feasible flip actions; flip history is provided as state information but is not a hard action mask.

The spin vector represents the current partition. Flipping the selected vertex updates the partition, cut value, and flip history. Search terminates at the prescribed step budget and returns the best cut encountered. ECO-DQN supplies preferences over the same vertex-flip actions, so no action-space aggregation is needed.

\section{Representative Generated Heuristics}
\label{app:generated_heuristics}

The following JSSP programs illustrate the decision rules generated by teacher-aware evolution and the performance-only EoH baseline.

\definecolor{codebg}{HTML}{F7F8FA}
\definecolor{codeframe}{HTML}{D9DEE7}
\definecolor{codekeyword}{HTML}{0033B3}
\definecolor{codestring}{HTML}{067D17}
\definecolor{codecomment}{HTML}{8C8C8C}
\definecolor{codenumber}{HTML}{9AA0A6}
\definecolor{codeidentifier}{HTML}{2B2B2B}

\lstdefinestyle{prettyPython}{
    language=Python,
    basicstyle=\ttfamily\scriptsize,
    keywordstyle=\color{codekeyword}\bfseries,
    stringstyle=\color{codestring},
    commentstyle=\color{codecomment}\itshape,
    identifierstyle=\color{codeidentifier},
    numbers=left,
    numberstyle=\tiny\color{codenumber},
    stepnumber=1,
    numbersep=8pt,
    showstringspaces=false,
    breaklines=true,
    breakatwhitespace=false,
    columns=fullflexible,
    keepspaces=true,
    tabsize=4,
    upquote=true,
    xleftmargin=0.5em,
    xrightmargin=0.5em,
}

\newtcblisting{pycode}[1][]{
    enhanced,
    breakable,
    colback=codebg,
    colframe=codeframe,
    boxrule=0.5pt,
    arc=2mm,
    left=1mm,
    right=1mm,
    top=1mm,
    bottom=1mm,
    listing only,
    listing options={style=prettyPython},
    colbacktitle=black,
    coltitle=white,
    fonttitle=\ttfamily\footnotesize,
    boxed title style={
        colback=black,
        colframe=black,
        boxrule=0pt,
    },
    #1
}

\label{app:code_jssp}

\begin{pycode}[title={JSSP heuristic generated by our method}]
# Task: JSSP
# Method: ours
# Algorithm:
# Retain the same critical-job backbone and progress-faded sync/bottleneck/risk interplay, but sharpen early-phase teacher-like wait absorption by strengthening the overlap gate, slightly softening machine-load pressure, and making remaining-work/lower-bound penalties more decisive with cleaner phase boundaries and tie-break bias toward less-congested, later-starting candidates.

def score(feature, state):
    flow_due_date = sum(state.instance.durations[feature.job_id][: feature.op_index + 1])

    processing_time = max(1.0, float(feature.processing_time))
    remaining_work = max(1.0, float(feature.remaining_work))
    machine_load = float(feature.machine_total_work)
    machine_queue = float(feature.machine_queue_len)
    earliest_start = float(feature.earliest_start)
    earliest_finish = float(feature.earliest_finish)
    machine_ready_time = float(feature.machine_ready_time)
    lower_bound_after = float(feature.lower_bound_after)

    total_ops = len(state.instance.durations[feature.job_id])
    progress = float(feature.job_progress)
    if progress <= 0.0 and total_ops > 0:
        progress = float(feature.op_index) / total_ops
    progress = min(1.0, max(0.0, progress))

    wait_for_machine = max(0.0, machine_ready_time - earliest_start)

    tail_ratio = remaining_work / (remaining_work + 0.72 * processing_time)
    denom_exp = 1.16 - 0.18 * progress
    criticality = -float(flow_due_date) / (remaining_work ** denom_exp)

    occupancy = processing_time / (1.0 + 0.62 * wait_for_machine)
    overlap = min(1.42, occupancy) * (0.72 * wait_for_machine + 0.22 * processing_time)
    substance_gate = 0.36 + 1.02 * (processing_time / (processing_time + 0.40 * remaining_work))
    phase_gate = max(0.0, 1.05 - 0.95 * progress)
    sync_term = phase_gate * substance_gate * overlap * (0.82 + 0.46 * tail_ratio)

    bottleneck_term = (0.00010 * machine_load + 0.00008 * machine_queue) * (0.68 + 0.58 * tail_ratio)

    risk_term = 0.00102 * earliest_finish + 0.00046 * lower_bound_after + 0.00006 * earliest_start

    excess_congestion = (machine_load / (1.0 + machine_load)) * (1.0 / (1.0 + 0.52 * wait_for_machine))
    congestion_penalty = 0.15 * (1.0 - phase_gate) * excess_congestion + 0.06 * phase_gate * excess_congestion

    tie_bias = 0.00003 * earliest_start - 0.00002 * machine_queue

    priority_score = (
        criticality
        + 0.0027 * sync_term
        + bottleneck_term
        - risk_term
        - congestion_penalty
        + tie_bias
    )
    return priority_score
\end{pycode}

\begin{pycode}[title={JSSP heuristic generated by EOH}]
# Task: JSSP
# Method: EoH
# Algorithm:
# Blend FDD/MWKR urgency with bottleneck pressure through a multiplicative criticality gate, then subtract a start-time slack penalty, so the rule promotes urgent low-remaining-work operations most strongly when they also sit on congested machines and can start soon.

def score(feature, state):
    import math

    durations = state.instance.durations[feature.job_id]
    flow_due_date = sum(durations[: feature.op_index + 1])

    remaining_work = max(1.0, float(feature.remaining_work))
    processing_time = float(feature.processing_time)
    machine_load = float(feature.machine_queue_len) + float(feature.machine_total_work)

    # Parent-2 core: urgency normalized by remaining work
    base_urgency = -flow_due_date / remaining_work

    # Parent-1 ingredients reorganized into a criticality gate
    # Stronger amplification when little work remains, machine is busy, and op is short.
    scarcity_gate = 1.0 + 1.0 / (remaining_work ** 0.7)
    congestion_gate = 1.0 + math.log1p(machine_load)
    quick_release_gate = 1.0 / (1.0 + processing_time)

    criticality = scarcity_gate * (0.65 + 0.35 * congestion_gate) * (0.8 + 0.2 * quick_release_gate)

    # New structural element: penalize delayed-start candidates relative to current frontier
    current_frontier = max(
        float(state.makespan),
        float(feature.job_ready_time),
        float(feature.machine_ready_time),
    )
    start_slack = max(0.0, float(feature.earliest_start) - current_frontier)
    slack_penalty = 0.12 * start_slack + 0.03 * max(0.0, float(feature.earliest_finish) - current_frontier)

    priority_score = base_urgency * criticality - slack_penalty
    return priority_score
\end{pycode}


\newpage
\section*{NeurIPS Paper Checklist}

\begin{enumerate}

\item {\bf Claims}
    \item[] Question: Do the main claims made in the abstract and introduction accurately reflect the paper's contributions and scope?
    \item[] Answer: \answerYes{} 
    \item[] Justification: The abstract and introduction describe teacher-guided heuristic evolution, its action-level feedback, and evaluation across scheduling, routing, and graph optimization.

\item {\bf Limitations}
    \item[] Question: Does the paper discuss the limitations of the work performed by the authors?
    \item[] Answer: \answerYes{} 
    \item[] Justification: Section~\ref{sec:conclusion} discusses dependence on teacher quality, compatible action interfaces, task-specific adapters, and analyzer calls during search.

\item {\bf Theory assumptions and proofs}
    \item[] Question: For each theoretical result, does the paper provide the full set of assumptions and a complete (and correct) proof?
    \item[] Answer: \answerNA{} 
    \item[] Justification:  This paper does not include theoretical results.

    \item {\bf Experimental result reproducibility}
    \item[] Question: Does the paper fully disclose all the information needed to reproduce the main experimental results of the paper to the extent that it affects the main claims and/or conclusions of the paper (regardless of whether the code and data are provided or not)?
    \item[] Answer: \answerYes{}
    \item[] Justification: Section~\ref{sec:method} and the appendices specify the algorithm, task interfaces, prompts, search budgets, and evaluation settings. The accompanying code provides configurations and reproduction instructions.

\item {\bf Open access to data and code}
    \item[] Question: Does the paper provide open access to the data and code, with sufficient instructions to faithfully reproduce the main experimental results, as described in supplemental material?
    \item[] Answer: \answerYes{}
    \item[] Justification: The implementation, generated programs, evaluation scripts, and data-access instructions are publicly available in the GitHub repository linked in the abstract.

\item {\bf Experimental setting/details}
    \item[] Question: Does the paper specify all the training and test details (e.g., data splits, hyperparameters, how they were chosen, type of optimizer) necessary to understand the results?
    \item[] Answer: \answerYes{}
    \item[] Justification: Section~\ref{sec:experimental_setup} and Appendix~\ref{app:configuration_provenance} specify the benchmark settings, evaluation budget, population size, and search allocations.

\item {\bf Experiment statistical significance}
    \item[] Question: Does the paper report error bars suitably and correctly defined or other appropriate information about the statistical significance of the experiments?
    \item[] Answer: \answerYes{}
    \item[] Justification: The experimental tables report means and sample standard deviations across generated programs.

\item {\bf Experiments compute resources}
    \item[] Question: For each experiment, does the paper provide sufficient information on the computer resources (type of compute workers, memory, time of execution) needed to reproduce the experiments?
    \item[] Answer: \answerYes{}
    \item[] Justification: Appendix~\ref{app:runtime_settings} reports CPU and GPU specifications, memory, software, and approximate search times and compute usage; the experimental tables report evaluation running times.
    
\item {\bf Code of ethics}
    \item[] Question: Does the research conducted in the paper conform, in every respect, with the NeurIPS Code of Ethics \url{https://neurips.cc/public/EthicsGuidelines}?
    \item[] Answer: \answerYes{}
    \item[] Justification: The research conducted in the paper conforms to the NeurIPS Code of Ethics.

\item {\bf Broader impacts}
    \item[] Question: Does the paper discuss both potential positive societal impacts and negative societal impacts of the work performed?
    \item[] Answer: \answerYes{}
    \item[] Justification: Appendix~\ref{app:broader_impacts} discusses inspectable deployment, risks from incomplete objectives, code defects and distribution shift, and validation measures.
    
\item {\bf Safeguards}
    \item[] Question: Does the paper describe safeguards that have been put in place for responsible release of data or models that have a high risk for misuse (e.g., pre-trained language models, image generators, or scraped datasets)?
    \item[] Answer: \answerNA{} 
    \item[] Justification: This work does not release high-risk pretrained models or scraped datasets.

\item {\bf Licenses for existing assets}
    \item[] Question: Are the creators or original owners of assets (e.g., code, data, models), used in the paper, properly credited and are the license and terms of use explicitly mentioned and properly respected?
    \item[] Answer: \answerYes{}
    \item[] Justification: We credit the original sources and follow the licenses and terms of use of the datasets, code, and pretrained models. Asset-specific license information accompanies the code.

\item {\bf New assets}
    \item[] Question: Are new assets introduced in the paper well documented and is the documentation provided alongside the assets?
    \item[] Answer: \answerYes{}
    \item[] Justification: The generated programs and implementation are documented with environment setup, usage examples, and evaluation instructions.

\item {\bf Crowdsourcing and research with human subjects}
    \item[] Question: For crowdsourcing experiments and research with human subjects, does the paper include the full text of instructions given to participants and screenshots, if applicable, as well as details about compensation (if any)? 
    \item[] Answer: \answerNA{} 
    \item[] Justification: This paper does not involve crowdsourcing nor research with human subjects.

\item {\bf Institutional review board (IRB) approvals or equivalent for research with human subjects}
    \item[] Question: Does the paper describe potential risks incurred by study participants, whether such risks were disclosed to the subjects, and whether Institutional Review Board (IRB) approvals (or an equivalent approval/review based on the requirements of your country or institution) were obtained?
    \item[] Answer: \answerNA{} 
    \item[] Justification: This paper does not involve crowdsourcing nor research with human subjects.

\item {\bf Declaration of LLM usage}
    \item[] Question: Does the paper describe the usage of LLMs if it is an important, original, or non-standard component of the core methods in this research? Note that if the LLM is used only for writing, editing, or formatting purposes and does \emph{not} impact the core methodology, scientific rigor, or originality of the research, declaration is not required.
    \item[] Answer: \answerYes{}
    \item[] Justification: LLMs analyze behavioral feedback and generate or revise heuristic code as core parts of Section~\ref{sec:method}. Settings and interface prompts appear in Section~\ref{sec:experimental_setup} and Appendix~\ref{app:task_interface_prompts}.

\end{enumerate}


\end{document}